\pdfoutput=1

\documentclass[11pt]{article}

\usepackage{acl}

\usepackage{times}
\usepackage{latexsym}

\usepackage[T1]{fontenc}

\usepackage[utf8x]{inputenc}

\usepackage{microtype}

\usepackage{array} %
\usepackage{setspace}
\usepackage{wrapfig}
\usepackage{times}
\usepackage{latexsym}
\usepackage{adjustbox} %
\usepackage{booktabs} %
\usepackage{multirow} %
\usepackage{hyperref}
\usepackage{microtype}
\usepackage{amssymb}
\usepackage{xspace}
\usepackage[shortlabels]{enumitem}
\usepackage{mathtools}
\usepackage{amssymb}
\usepackage{footnote}
\usepackage{tablefootnote}
\usepackage{graphicx}
\usepackage{color, colortbl}
\usepackage{xstring}
\usepackage{comment}
\usepackage{float}
\restylefloat{table}
\usepackage{array,booktabs,makecell}
\usepackage{appendix}
\usepackage[normalem]{ulem}
\usepackage{arydshln}
\usepackage{scalerel,xparse}
\usepackage{subcaption}
\usepackage{wrapfig,lipsum,booktabs}
\usepackage{titlesec}

\newcolumntype{H}{>{\setbox0=\hbox\bgroup}c<{\egroup}@{}}

\newcolumntype{L}{>{\centering\arraybackslash}m{1.2cm}} %
\newcolumntype{R}{>{\centering\arraybackslash}m{0.6cm}} %
\newrobustcmd{\B}{\bfseries} %
\usepackage{siunitx} %
\newcommand{\savefootnote}[2]{\footnote{\label{#1}#2}}
\newcommand{\repeatfootnote}[1]{\textsuperscript{\ref{#1}}}

\usepackage[autostyle]{csquotes}

\title{Improving Neural Machine Translation of Indigenous Languages with Multilingual Transfer Learning}%

\author{Wei-Rui Chen ~~~~~~~~  Muhammad Abdul-Mageed \\
\normalsize Natural Language Processing Lab  \\
  \normalsize The University of British Columbia\\
  \texttt{ \small \{weirui.chen,muhammad.mageed\}@ubc.ca}
  }

\begin{document}

\maketitle

\begin{abstract}
Machine translation (MT) involving Indigenous languages, including those possibly endangered, is challenging due to lack of sufficient parallel data. We describe an approach exploiting bilingual and multilingual pretrained MT models in a transfer learning setting to translate from Spanish to ten South American Indigenous languages. Our models set new SOTA on five out of the ten language pairs we consider, even doubling performance on one of these five pairs. Unlike previous SOTA that perform data augmentation to enlarge the train sets, we retain the low-resource setting to test the effectiveness of our models under such a constraint. In spite of the rarity of linguistic information available about the Indigenous languages, we offer a number of quantitative and qualitative analyses (e.g., as to morphology, tokenization, and orthography) to contextualize our results.

\end{abstract}

\section{Introduction}
\begin{figure*}[h]
\begin{centering}
\includegraphics[scale=0.36]{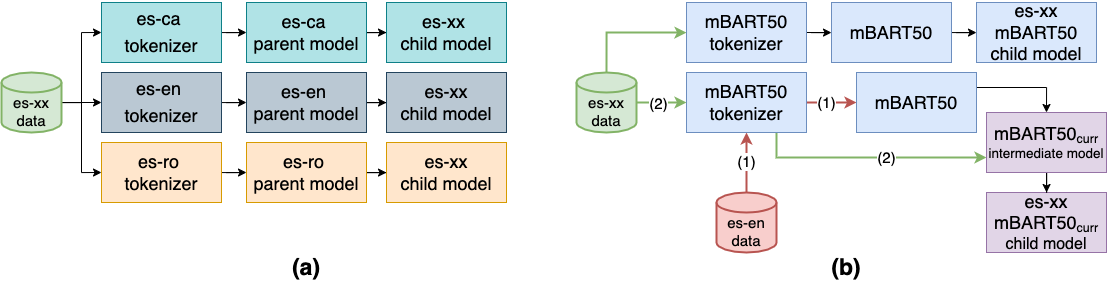}
  \caption{Model Training in \textbf{(a)} bilingual setting and \textbf{(b)} multilingual setting for one \texttt{es-xx} language pair. For both \textbf{(a)} and \textbf{(b)}, child models are those being used for prediction. xx represents arbitrary one of the ten South American Indigenous languages. For \textbf{(b)}, the blue es-xx mBART50 child model represents the model directly fine-tuned with \texttt{es-xx} data. The purple es-xx mBART50\textsubscript{curr} child model represents the model that is first being fine-tuned with \texttt{es-en} data to produce an intermediate model, indicated as (1). Afterwards, it is fine-tuned with \texttt{es-xx} data, indicated as (2).}
  \label{fig:flowchart}
 \end{centering}
\end{figure*}

Artificial intelligence (AI) is being widely integrated into many natural language processing (NLP) applications in our daily lives. However, these language technologies have focused almost exclusively on widely-spoken languages~\cite{choudhury2021linguistically}. Under-represented languages such as endangered languages are thus left out. For example, the Google machine translation (MT) system does not support any of the languages included in this study.\footnote{https://translate.google.com/about/languages/} Our objective in this work is hence to build machine translation (MT) models for Indigenous languages, which are by definition low-resource and possibly endangered. More specifically, we focus on South American Indigenous languages. In a MT scenario, a language pair is considered `low-resource' if the parallel corpora consists of less than $0.5$ million of parallel sentences and `extremely low-resource' if less than $0.1$ million of parallel sentences~\cite{ranathunga2021neural}. In this work, nine out of ten languages pairs we consider have under $0.1$ million pairs of sentences (with only one language pair having roughly $0.1$ million pairs of sentences). Developing MT systems for endangered  languages can help preserve these languages. MT systems, for example, can enable translating abundant textual resources written in major world languages into endangered languages. 

Deep learning opened up a new era of AI~\cite{lecun2015deep}, making artificial neural networks (ANNs) one of the most important areas of machine learning. Neural Machine Translation (NMT) is a subfield of MT that leverages ANNs to build translation systems. NMT has become in recent years the leading method for carrying our MT. With developments in deep learning methods and technologies - new model architectures and training strategies - NMT keeps surpassing the best performances set previously by itself~\cite{barrault-etal-2020-findings,barrault-etal-2019-findings}. Despite that NMT is able to produce powerful MT systems, it is data-hungry. That is, it requires large amounts of data to train a quality NMT model~\cite{koehn-knowles-2017-six}. Contemporary machine translation systems are oftentimes trained on over a million of parallel sentences~\cite{fan2021beyond,tang2020multilingual} for high-resource language pairs. In contrast, the size of the dataset we have is limited. Transfer learning has been shown to help mitigate this issue by porting knowledge e.g. from a parent model to a child model~\cite{zoph-etal-2016-transfer}. Therefore, we investigate the utility of transfer learning and NMT on low-resource South American Indigenous languages.

Namely, we offer NMT models in the context of transfer learning that translate from Spanish into $10$ South American Indigenous languages. We leverage two types of pretrained MT models: \textit{bilingual} models and a \textit{multilingual} model. The overall training approach is illustrated in Figure~\ref{fig:flowchart}. Our downstream datasets are provided by AmericasNLP2021~\cite{mager-etal-2021-findings} shared task. We compare our performance to the winner of the shared task~\cite{vazquez-etal-2021-helsinki}.

The rest of this study is organized as follows: Section~\ref{sec:background} is a literature review on Indigenous MT, transfer learning, the application of transfer learning to NMT, and the challenge of cross-lingual transfer. In Section~\ref{sec:method}, we describe our experimental settings where we provide information about datasets, baselines, data preprocessing, our bilingual and multilingual models, and our training procedures. We present our results in Section~\ref{sec:results}, and provide discussions in Section~\ref{sec:discussion}. We conclude in Section~\ref{sec:conclusion}.
\section{Background} \label{sec:background}

\subsection{MT on Indigenous Languages}\label{low_resource_indigenous}

Languages are diverse. For example, in South America, there are $108$ language families, $55$ of which are in a language family with one single member (i.e., language isolates)~\cite{campbell2012indigenous}. Due to this linguistic diversity, to the best of our knowledge, there is no single MT method that fits all Indigenous languages. However, since many Indigenous languages suffer the low-resource issue~\cite{mager-etal-2018-challenges}, many researchers borrow ideas from low-resource MT to tackle the task of MT of Indigenous languages. We survey some approaches here.

~\citet{DBLP:journals/corr/abs-2104-07483} create models based on the T5 architecture ~\cite{DBLP:journals/corr/abs-1910-10683} and train it with monolingual Indigenous data before fine-tuning on parallel data, thus attempting to acquire knowledge of the Indigenous languages to benefit MT.~\citet{vazquez-etal-2021-helsinki} augment the parallel data by collecting monolingual Indigenous data and translate  them to Spanish. They then train a multilingual MT model with the augmented parallel data.~\citet{ngoc-le-sadat-2020-revitalization} focus on data preprocessing, and build a morphological segmenter for the source language Inuktitut to achieve better performance in Inuktitut-English translation. These aforementioned works all adopt methods invented to tackle the task of MT on low-resource languages~\cite{ranathunga2021neural,DBLP:journals/corr/abs-2109-00486}. We now introduce transfer learning, which is where our approach falls.

\subsection{Transfer Learning and NMT}
It can sometimes be very expensive to collect data for MT. This is true especially for endangered languages when the number of speakers is decreasing. Therefore, many endangered languages suffer from the the low-resource issue. This motivates methods that can help port knowledge from existing resources to a down-stream task of interest with low-resources employing transfer learning methods. An additional motivation for studying and applying transfer learning is that human beings are able to apply knowledge/skills they acquired earlier from some jobs to better perform new related jobs with less efforts. An analogy is this: a person who has learned a music instrument may be able to pick up another instrument easier and quicker~\cite{zhuang2020comprehensive}. When applying transfer learning in the context of NMT, a scenario can be as follows: a model previously trained on parent language pair(s) (called \textit{parent model}) is further fine-tuned on child language pair(s) to form a \textit{child model}. Under such a scenario, a parent language pair is one of the language pairs whose bilingual data is used to train a model from scratch and produce a parent model. A child language pair is one of the language pairs whose bilingual data is used to fine-tune a parent model and produce a child model. Again, the intuition here is that an experienced translator (pretrained MT model) on one language pair may be able to translate into another language pair with shorter time and less effort compared to a unexperienced person (new randomly-initialized model). The core idea is to retain the parameters of parent model as the starting point for the child model, instead of training from scratch where the parameters are randomly initialized~\cite{zoph-etal-2016-transfer, DBLP:journals/corr/abs-1809-00357, DBLP:journals/corr/abs-1708-09803}.

\subsection{Cross-lingual Transfer}\label{cross_ling_transfer}
One of the challenges of transfer learning in MT is the mismatch in parent and child vocabularies. Only when the parent language pair and child language pair are identical can there be no such issue. Otherwise, when at least one of the languages in child language pair is distinct from parent languages, such an issue would arise. This is the case since vocabulary is language-specific and discrete~\cite{kim-etal-2019-effective}. For example, if a parent model has its vocabulary built upon Spanish-English text, the vocabulary will contain only Spanish and English tokens. It can be unpredictable when tokenizing French text with such a vocabulary. 

\citet{zoph2016transfer} tackle this challenge by retaining the token embeddings for their target language since the parent target language and child target language are the same in their work. For parent and child source languages, they randomly map tokens of parent source language to tokens of child source language. \citet{DBLP:journals/corr/abs-1809-00357} take another approach of vocabulary building: the vocabulary is built upon $50\%$ of parallel sentences of the parent language pair and $50\%$ of those of the child language pair, so the vocabulary will contain tokens of both parent and child language pairs.~\citet{kocmi-bojar-2020-efficiently} introduce yet another simpler idea named `Direct Transfer' where the parent vocabulary is used to train a child model. Although the parent vocabulary is not optimized for child language pair and can oversegment words in child language pair to smaller pieces than necessary, such a method still shows significant improvement in many language pairs.~\citet{kocmi-bojar-2020-efficiently} suspect that this could be due to good generalization of the transformer architecture to short subwords.

\section{Experiments}\label{sec:method}

\begin{figure}[t]
\begin{centering}
\includegraphics[scale=0.15]{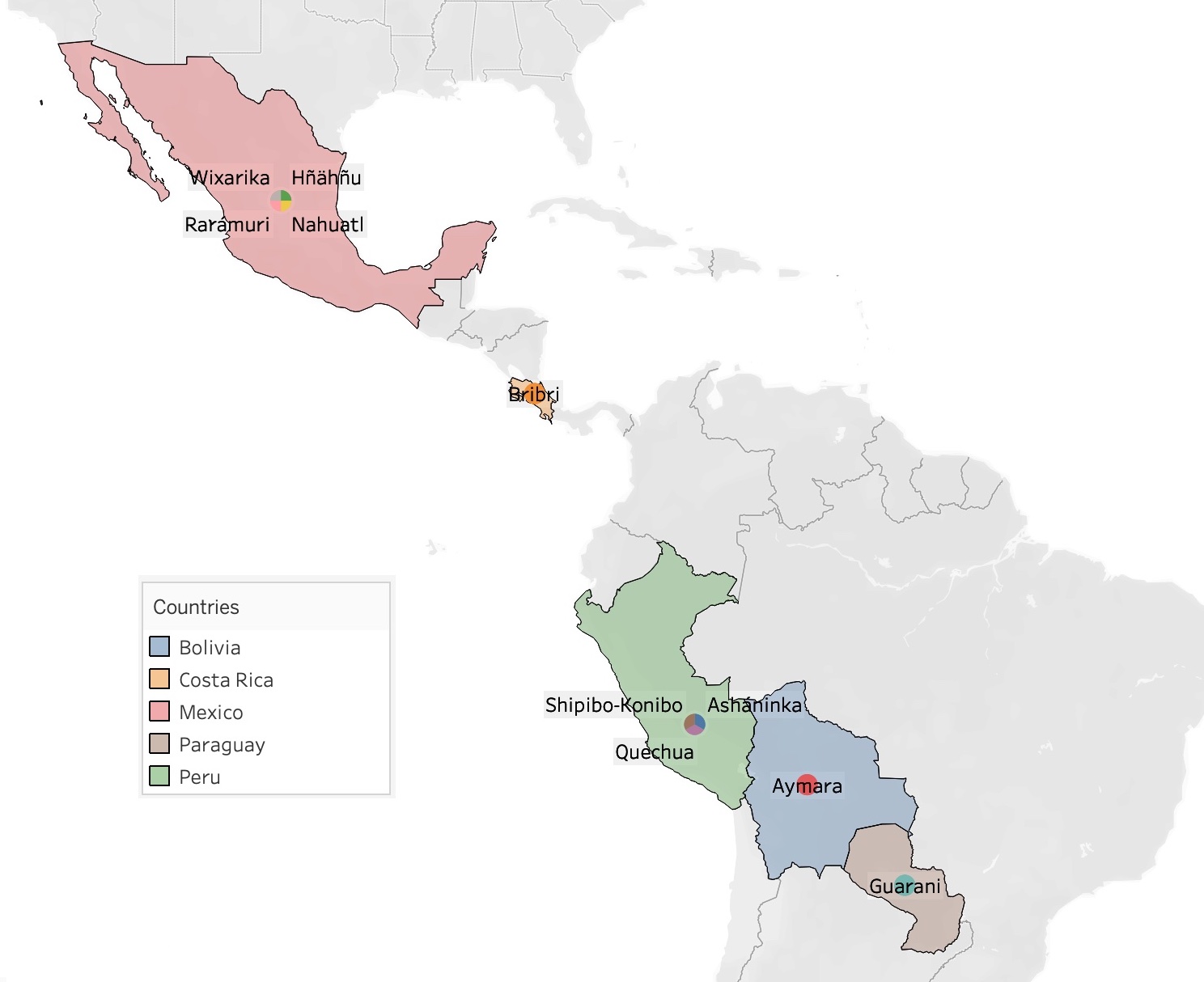}
  \caption{A map of the ten South American Indigenous languages in our data. The color for each country and each language is arbitrarily assigned.}
  \label{fig:map}
 \end{centering}
\end{figure}

\begin{table}[t]
\footnotesize
\centering
 \begin{tabular}{llcr}
 \toprule
\textbf{Language} & \textbf{ISO}  & \textbf{Major location}  & \textbf{Speakers} \\
\toprule
Aymara          & aym  & Bolivia   & 1,677,100 \\ 
Bribri          & bzd  & Costa Rica   & 7,000   \\
Asháninka       & cni  & Peru  & 35,200          \\ 
Guarani         & gn   & Paraguay & 6,652,790   \\
Wixarika        & hch  & Mexico   & 52,500  \\
Nahuatl         & nah  & Mexico   & 410,000   \\
Hñähñu          & oto  & Mexico   & 88,500 \\
Quechua         & quy  & Peru   & 7,384,920   \\
Shipibo-Konibo  & shp  & Peru   &  22,500   \\
Rarámuri        & tar  & Mexico   & 9,230   \\

\toprule 
\end{tabular}
\caption{Overview of the ten Indigenous languages ~\cite{Ethnologue2021}.}  \label{tab:langs_overview}

\end{table}
\normalsize
\begin{table}[h]
\centering
\footnotesize
 \begin{tabular}[t]{lrcc}
 \toprule
\textbf{Language Pair} & \textbf{Train}  & \textbf{Dev}  & \textbf{Test} \\
\toprule
es-aym    & $6,531$ & $996$   & $1,003$   \\
es-bzd   & $7,506$  & $996$   & $1,003$   \\
es-cni  & $3,883$  & $883$   & $1,003$ \\
es-gn   & $26,032$ & $995$   &  $1,003$   \\
es-hch  & $8,966$ & $994$   & $1,003$  \\
es-nah    & $16,145$ & $672$   & $996$   \\
es-oto   & $4,889$ & $599$   & $1,001$ \\
es-quy  & $125,008$  & $996$   & $1,003$   \\
es-shp  & $14,592$ & $996$   &  $1,003$   \\
es-tar   & $14,720$  & $995$   & $1,003$   \\

\toprule 
\end{tabular}
\caption{Number of parallel sentences}  \label{tab:dataset_split_datapoints}
\end{table}
\normalsize

\subsection{Dataset}\label{sec:datasets}
Our dataset is from AmericasNLP 2021 Shared Task on Open Machine Translation, which was co-located with the 2021 Annual Conference of the North American Chapter of the Association for Computational Linguistics (NAACL-HLT 2021)~\cite{mager-etal-2021-findings}. The dataset contains  parallel data of $10$ language pairs: from Spanish to Aymara, Asháninka, Bribri, Guarani, Hñähñu, Nahuatl, Quechua, Rarámuri, Shipibo-Konibo, and Wixarika. An overview of these $10$ Indigenous languages is shown in Table~\ref{tab:langs_overview}. The geographical distribution of the languages is depicted in Figure~\ref{fig:map}. We offer information about the dataset splits as distributed by the shared task organizers in Table~\ref{tab:dataset_split_datapoints}. The shared task has two tracks: \textbf{Track One}, where the training split (Train) involves an arbitrary portion of development set, and \textbf{Track Two}, where Train involves \textit{no} development data. In this work, we take \textit{Track One} as our main focus and concatenate $90\%$ of Dev split to Train to acquire a bigger training set. We also conduct experiments for \textit{Track Two}, and we put the results in Appendix.

\begin{table*}[t]
\small
\centering
\begin{adjustbox}{width=15cm}
\renewcommand{\arraystretch}{1.5}
{
        \begin{tabular}{lHll}
        \toprule

      \textbf { Pair }  & \textbf{Type} & \textbf {~~~~~~~~~~~~~~~~~~~~~~~~~~~~~~~~~Source} & \textbf{~~~~~~~~~~~~~~~~~~~~~~~~~~~~~~~~~Target} \\    \toprule
            
\multirow{2}{*}{\textbf{es-aym}} & original &\colorbox{blue!8}{ Los artistas de IRT ayudan a los niños en las escuelas.}   &      \small{ \colorbox{blue!8}{IRT artistanakax jisk'a yatiqañ utankir wawanakaruw yanapapxi.}}\\ 
  & tokenized & \colorbox{green!10}{Los artistas de I RT ayudan a los niños en las escuelas .}  &   \colorbox{green!10}{ \small{ I RT artist ana ka x  ji sk ' a y ati qa ñ u tank ir wa wan aka ru w ya nap ap xi . } }\\ 
\hline

\multirow{2}{*}{\textbf{es-bzd}   } & original &\colorbox{blue!8}{ Fui a un seminario que se hizo vía satélite.   }   &    \colorbox{blue!8}{  \small{Ye' dë'rö seminario ã wéx yö' satélite kĩ.}}\\ 
  & tokenized & \colorbox{green!10}{Fui a un seminario que se hizo vía satélite .}   &    \colorbox{green!10}{  \small{Ye ' d ë ' r ö seminar io  ã w é x y ö ' sat éli te k ĩ .} }\\ 
\hline

 \multirow{2}{*}{\textbf{es-cni}   } & original  &  \colorbox{blue!8}{Pensé que habías ido al campamento.} & \colorbox{blue!8}{\small{Nokenkeshireashitaka pijaiti imabeyetinta.}}\\
  & tokenized &  \colorbox{green!10}{Pensé que había s ido al campamento.} & \colorbox{green!10}{\small{No ken ke shire ashi t aka p ija iti im ab eye tin ta .}}\\

\hline

  \multirow{2}{*}{\textbf{es-gn}   }& original &  \colorbox{blue!8}{Veía a su hermana todos los días.}     &      \colorbox{blue!8}{\small{Ko'êko'êre ohecha heindýpe.}}\\
 & tokenized  &   \colorbox{green!10}{Ve ía a su hermana todos los días .}        &      \colorbox{green!10}{\small{Ko ' ê ko ' ê re oh e cha he in d ý pe .}}\\
\hline

\multirow{2}{*}{\textbf{es-hch}   } & original &  \colorbox{blue!8}{Era una selva tropical.} &      \colorbox{blue!8}{\small{pe h+k+t+kai metsi+ra+ ye tsie nieka ti+x+kat+.  }}\\
 & tokenized  &  \colorbox{green!10}{Era una selva tropical .}  &      \colorbox{green!10}{\small{pe h + k + t + ka i met si + ra + ye t s ie  nie ka ti + x + ka t + .}}\\
\hline

 \multirow{2}{*}{\textbf{es-nah}   } & original  &   \colorbox{blue!8}{Santo trabajó para Disney y operó las tazas de té.}     &      \colorbox{blue!8}{\small{zanto quitequitilih Disney huan quinpexontih in cafen caxitl   }}\\
 & tokenized &  \colorbox{green!10}{Santo trabajó para Disney y o per ó las taza s de té .}  &      \colorbox{green!10}{\small{zan to quite qui til ih Disney h uan quin pex on t ih in cafe n ca xi t l}}\\
\hline

\multirow{2}{*}{\textbf{es-oto}   } & original &\colorbox{blue!8}{ Otros continúan reconociendo nuestro éxito.} &      \colorbox{blue!8}{\small{ymana ditantho anumahditho goma npâgu}}\\
 & tokenized  &   \colorbox{green!10}{Otros continúan reconociendo nuestro éxito . }   &      \colorbox{green!10}{\small{y man a di tant ho an um ah di th o go ma n p â gu}}\\
\hline

\multirow{2}{*}{\textbf{es-quy}   } & original &  \colorbox{blue!8}{De vez en cuando me gusta comer ensalada. }   &      \colorbox{blue!8}{\small{Yananpiqa ensaladatam mikuytam munani  }}\\
 & tokenized  &  \colorbox{green!10}{De vez en cuando me gusta comer ensalada . } &      \colorbox{green!10}{\small{Yan an pi qa en s ala data m m iku y tam  mun ani }}\\
\hline

\multirow{2}{*}{\textbf{es-shp}   } & original &  \colorbox{blue!8}{El Museo se ve afectado por las inversiones. }  &      \colorbox{blue!8}{\small{Ja Museora en oinai inversionesbaon afectana.  }}\\
 & tokenized  &  \colorbox{green!10}{ El Museo se ve afectado por las inversiones . } &      \colorbox{green!10}{\small{Ja Museo ra en o ina i in version es ba on a fect ana .  }}\\
\hline
 
  \multirow{2}{*}{\textbf{es-tar}   } & original  &  \colorbox{blue!8}{Es un hombre griego.} &      \colorbox{blue!8}{\small{Bilé rejói Griego ju }}\\
 & tokenized &   \colorbox{green!10}{Es un hombre griego .} &      \colorbox{green!10}{\small{Bil é re j ó i Gri ego ju}}\\

\hline
\toprule

\end{tabular}}
\end{adjustbox}
        \caption{Example sentences tokenized by \texttt{es-en} tokenizer. \colorbox{blue!8}{\textbf{Light blue}}: Original sentences (source or target).~\colorbox{green!10}{\textbf{Light green}}:~tokenized sentenses with tokens separated by whitespace. }
    
    \label{tab:tokenization-result}
\end{table*}

\subsection{Baselines}\label{sec:baselines} %

We compare our results with the winner of the shared task~\citet{vazquez-etal-2021-helsinki} who achieve highest performance in evaluation metrics for all language pairs in Track One (and winning $9$ out of $10$ language pairs in Track Two). They augment the training data by (1) gathering external parallel data, e.g. Bibles and Constitutions (2) collecting monolingual data of Indigenous languages and adopt back-translation method to generate synthetic parallel data. They build a 6-layered transformer~\cite{vaswani2017attention} with 8 heads by first pretrain it with \texttt{es-en} parallel data and then fine-tune it with both internal dataset provided by the  organizer and external augmented datasets of all $10$ language pairs to produce a multilingual MT model. In this work, we leverage solely the dataset provided by the shared task organizer to test if the method works with scarce data.

\subsection{Data Preprocessing}\label{data_preprocessing}
As mentioned in section~\ref{cross_ling_transfer}, the cross-lingual challenge exists  when one or both sides of child language pair is distinct from the parent languages which is the case to all of the our 10 language pairs. To tackle this, we opt for `direct transfer' method, due to its simplicity, to exploit parent vocabulary for child model. As ~\citet{kocmi-bojar-2020-efficiently} find that the words of child language are oversegmented with direct transfer, similar to their finding, we observe that the words of Indigenous language words can be oversegmented. As shown in Table~\ref{tab:tokenization-result}, it can be seen that the source sentences are tokenized reasonably well with mostly one token per word. By contrast, the words of child target language are generally oversegmented into short subwords. The statistics of the tokenization is shown in Table~\ref{tab:token_stats_ninty_percent_dev}. An analysis of oversegmentation phenomenon is given in section~\ref{target_side_subword_issue}.

\subsection{Parent Models} %
We offer two types of parent models, bilingual models and multilingual models.

\noindent\textbf{Bilingual Models.} For bilingual models, we leverage publicly accessible pretrained models from Huggingface~\cite{wolf-etal-2020-transformers} as provided by Helsinki-NLP~\cite{TiedemannThottingal:EAMT2020}. The pretrained MT models released by Helsinki-NLP are trained on OPUS, an open source parallel corpus~\cite{tiedemann-2012-parallel}. Underlying these models is the Transformer architecture of Marian-NMT framework implementation~\cite{junczys-dowmunt-etal-2018-marian}. Each model has six self-attention layers in encoder and decoder parts, and each layer has eight attention heads. The three bilingual models we specifically use are each pretrained with OPUS Spanish-Catalan, Spanish-English, and Spanish-Romanian data.\footnote{\citet{TiedemannThottingal:EAMT2020} do not provide information about the size of OPUS data exploited in each of these models.}

We choose these models because their source language is Spanish so they may have good Spanish subword embeddings. In this regard, as~\citet{adelaar2012chapter} point out, during the colonial period, Spanish grammatical concepts were introduced to some South American Indigenous languages. In addition, we pick Spanish-Catalan and Spanish-Romanian MT models because Catalan and Romanian are two languages in the same Romance language family as Spanish, and we suspect our ten Indigenous languages of South America may have some affinity to Spanish. We also choose Spanish-English as a contrastive model because English is in the Germanic language family rather than Romance and that the MT models built around English usually are well-performing due to its rich resource of parallel data.

\noindent\textbf{Multilingual Models.} For our multilingual models, we exploit mBART50~\cite{tang2020multilingual}.  mBART50 can be seen as an extension of mBART~\cite{liu-etal-2020-multilingual-denoising}. mBART (or more specifically mBART25)  is a multilingual sequence-to-sequence generative model pretrained on $25$ monolingual datasets and fine-tuned on 24 bilingual datasets which cover all $25$ languages used in pretraining. mBART50 takes mBART as a starting point and enlarges its embedding layers to accommodate tokens of $25$ new languages to support $50$ languages. mBART50 adopts multilingual fine-tuning under three scenarios: one-to-many, many-to-one, and many-to-many where `one' represents English. We choose the one that is trained under many-to-many scenario to ensure (1) Spanish is fine-tuned as a source language so it may maintain a good representation for Spanish tokens (2) \texttt{es-en} language pair is covered so we can produce an intermediate model with \texttt{es-en} fine-tuning to test the effectiveness of curriculum learning.

\subsection{Training Approach}

\noindent\textbf{Bilingual Model Training.} We fine-tune each of our three bilingual models for $60,000$ steps with Spanish-Indigenous data, acquiring performance on Dev every $1,000$ steps. The final model is the checkpoint that has the lowest validation/Dev loss, and it is what we use for predicting on Test. Our beam size (for beam search)~\cite{reddy1977speech,DBLP:journals/corr/abs-1211-3711} is $6$. We use a batch size~\savefootnote{batchSize}{The batch sizes are small so the data can be loaded in the GPU memory.} of $15$ for our bilingual models. It takes  $\sim6$ hours to train on four Nvidia V100-SXM2-16GB GPUs for each model per language pair.

\noindent\textbf{Multilingual Model Training.} For our multilingual setting, we train a model for each of the Spanish$\rightarrow$Indigenous language pairs and it takes $\sim12$ hours to train on four NVIDIA Tesla V100 32GB NVLink GPUs for each model per language pair. We have two scenarios:
mBART50 and mBART50\textsubscript{curr}. Both of them have batch size~\repeatfootnote{batchSize} to be $5$, and the beam size to be $6$.

\begin{table}[t]
\tiny 
\centering
 \begin{tabular}{
    c
    R
    S[table-number-alignment = right]
    S[table-number-alignment = right]
    R
    R
    }
 \toprule
\multicolumn{1}{c}{\textbf{Model}} &
\multicolumn{1}{R}{\textbf{Target}} &

\multicolumn{1}{c}{\textbf{Our BLEU}} & 
\multicolumn{1}{c}{\textbf{Our chrF}}  & 
\multicolumn{1}{R}{\textbf{SOTA BLEU}} & 
\multicolumn{1}{R}{\textbf{SOTA chrF}} \\
\toprule
es-ca &  & 1.445 & 0.2344  \\
es-en & \multirow{3}{*}{aym}  & 2.432 & 0.277 & \multirow{3}{*}{\B{2.8}} & \multirow{3}{*}{\B{0.31}} \\
es-ro &    & 2.009  & 0.2705 &  \\
mBart50 &  & 2.017 & 0.2672\\
mBART50\textsubscript{curr} &  & 2.23 & 0.2725\\
\hdashline
es-ca &   & 7.242 & 0.2378 &  & \\
es-en & \multirow{3}{*}{bzd}  & 9.952 & 0.2753 & \multirow{3}{*}{5.18} & \multirow{3}{*}{0.213}  \\
es-ro &   & 10.278 & 0.2867 & \\
mBart50 &  & \B 12.898 & \B 0.3082  \\
mBART50\textsubscript{curr} & & 12.495 &  0.3036  \\
\hdashline
es-ca &   & 4.742 & 0.2984 &  & \\
es-en & \multirow{3}{*}{cni}  & 5.973 & 0.3367 & \multirow{3}{*}{6.09} & \multirow{3}{*}{0.332}  \\
es-ro &    & 5.21  & 0.3229 & \\
mBart50 &   & 5.632 & 0.3183\\
mBART50\textsubscript{curr} &   & \B 6.255 & \B 0.3432\\
\hdashline
es-ca &   & 4.395 & 0.2909 &  &  \\
es-en & \multirow{3}{*}{gn}  & 5.918 & 0.3341 & \multirow{3}{*}{\B{8.92}} & \multirow{3}{*}{\B{0.376}}  \\
es-ro &   & 5.853 & 0.3279 & \\
mBart50 &   & 6.329 & 0.3367\\
mBART50\textsubscript{curr} &  & 6.449 & 0.3387\\
\hdashline
es-ca &   & 13.375 & 0.3061 &  &  \\
es-en & \multirow{3}{*}{hch}  & 15.922 & 0.3461 & \multirow{3}{*}{15.67} & \multirow{3}{*}{\B{0.36}}  \\
es-ro &   & 15.298 & 0.3444 & \\
mBart50 &  & \B 16.731 & 0.3397 &  \\
mBART50\textsubscript{curr} &  & 16.659 & 0.3391 &  \\
\hdashline
es-ca &   & 1.95 & 0.2763 &  &  \\
es-en & \multirow{3}{*}{nah}  & 2.045 & 0.2913 & \multirow{3}{*}{\B{3.25}} & \multirow{3}{*}{0.301}  \\
es-ro &   & 1.734 & 0.2929 &  \\
mBart50 &  & 2.422 & 0.2969 & \\
mBART50\textsubscript{curr} &  & 2.947 & \B 0.3015 & \\
\hdashline
es-ca &   & 4.344 & 0.2268 &  &  \\
es-en & \multirow{2}{*}{oto}  & 6.414 & 0.2522 & \multirow{3}{*}{5.59} & \multirow{3}{*}{0.228} \\
es-ro &   & 4.14 & 0.2315 &  \\
mBart50 &   & \B 7.504  & \B 0.265\\
mBART50\textsubscript{curr} &   & 7.489  & 0.2617\\
\hdashline
es-ca &   & 2.817 & 0.3449 &  &  \\
es-en & \multirow{3}{*}{quy}  & 4.149 & 0.3788 & \multirow{3}{*}{\B{5.38}} & \multirow{3}{*}{\B{0.394}} \\
es-ro &   & 3.192 & 0.3718 &   \\
mBart50 &   & 4.689  & 0.3928 \\
mBART50\textsubscript{curr} &   & 4.95  & 0.3881 \\
\hdashline
es-ca &   & 5.184 & 0.2627 &  &  \\
es-en & \multirow{3}{*}{shp}  & 7.664 & 0.3326 & \multirow{3}{*}{\B{10.49}} & \multirow{3}{*}{\B{0.399}}  \\
es-ro &   & 6.663 & 0.32 &  \\
mBart50 &   & 10.022  & 0.3556\\
mBART50\textsubscript{curr} &   & 9.702  & 0.349\\
\hdashline
es-ca &  & 1.724 & 0.217 &  &  \\
es-en & \multirow{3}{*}{tar}  & 2.432 & 0.248 & \multirow{3}{*}{\B{3.56}} & \multirow{3}{*}{\B{0.258}}  \\
es-ro &   & 2.034 & 0.2358 &  \\
mBart50 & & 2.433 & 0.2396 \\
mBART50\textsubscript{curr} &  & 2.261 & 0.2362 \\

\toprule 
\end{tabular}
\caption{Modeling results (of Track One). The boldfaced numeric values are the best performances. Source language is always Spanish so it is ignored. SOTA values represent the state-of-the-art performance which are all from~\citet{vazquez-etal-2021-helsinki}}  \label{tab:model_results_ninty_dev}

\end{table}
\normalsize
\begin{table*}[t]
\small
\centering
\begin{adjustbox}{width=14cm}
\renewcommand{\arraystretch}{1.6}
{
        \begin{tabular}{lHl}
        \toprule

      \textbf { Pair }  & \textbf{Type} &  \textbf{~~~~~~~~~~~~~~~~~~~~~~~~~~~~~~~~~~~~~~~~~~~~~~~~~~~~~~~~~~~~~~~~~~~~~~~~~~~~~~~~~~~~~~~~~Sentence} \\ \toprule
            
\multirow{2}{*}{\textbf{es-aym}} & Ground Truth & \colorbox{blue!6}{nanakan utaxax khaysa Concord uksanx kimsatunka waranqa acres ukhamarac walja uywanakarakiw utjaraki.   }\\ 
  & Prediction  &  \colorbox{green!10}{Concord markan nanakan utanx 30000 acre ukhamarak walja uywanaka utji.} \\ 
\hline

\multirow{2}{*}{\textbf{es-bzd}   } &  Ground Truth & \colorbox{blue!6}{Sa' ù Concord wã 30000 acres tã' ẽnã tãîx íyiwak.}\\ 
  & Prediction  & \colorbox{green!10}{Sa' ù ã Concord e' kĩ káx dör 20.000 acres tãîx íyiwak tãîx. }  \\ 
  
\hline

 \multirow{2}{*}{\textbf{es-cni}} & Ground Truth  &  \colorbox{blue!6}{Abanko Concordki otimi 30000 acres jeri osheki birantsipee.} \\ 
  & Prediction  & \colorbox{green!10}{Ashi pankotsi Concordi timatsi 30000 acres aisati osheki piratsipee. }\\ 
\hline

  \multirow{2}{*}{\textbf{es-gn}}& Ground Truth & \colorbox{blue!6}{ Ore róga Concord-pe otroko 30000 acres ha hetaiterei orerymba.}\\ 
 & Prediction   &  \colorbox{green!10}{Ñane óga Concord-pe oreko 30000 acre ha hetaiterei mymba.}\\
\hline

\multirow{2}{*}{\textbf{es-hch}   } & Ground Truth & \colorbox{blue!6}{ta kí wana Concord pe xeiya 30000 acres tsiere y+ wa+kawa yeuta meteu uwa.} \\ 
 & Prediction   & \colorbox{green!10}{ta ki wana Concord pexeiya xeiya xeitewiyari acre meta wa+kawa te+teri.  }\\ 
\hline

 \multirow{2}{*}{\textbf{es-nah}   } & Ground Truth  &  \colorbox{blue!6}{tochan Concord quipiya miyac tlalli nohiya miyac tlapiyalli.}\\ 
 & Prediction  & \colorbox{green!10}{Tehuancalco Concord quipiah macualli tlatqui ihuan miyac yolcameh.}\\ 
\hline

\multirow{2}{*}{\textbf{es-oto}   } & Ground Truth & \colorbox{blue!6}{mangû game ane Concord phodi 30000 yñi xi nā hmudi on yzuî} \\ 
 & Prediction   &  \colorbox{green!10}{Goma na madoongû ane Concord phodi 30000 yqhēya xi na ngû on ybaoni } \\
\hline

\multirow{2}{*}{\textbf{es-quy}   } & Ground Truth & \colorbox{blue!6}{Corcord nisqapi wasiykum kimsa chunka waranqa acres nisqan kan hinataq achkallaña uywa.} \\ 
 & Prediction  & \colorbox{green!10}{Concordpi wasiykuqa 30000 acres hinaspa achka uywakunam }\\ 
\hline

\multirow{2}{*}{\textbf{es-shp}   } & Ground Truth & \colorbox{blue!6}{Non xobo Concordainra 30000 acresya iki itan kikin icha yoinabo.}\\ 
 & Prediction  & \colorbox{green!10}{Concordainra non xoboa riki 30000 acres itan kikin icha yoinabo. } \\ 
\hline
 
  \multirow{2}{*}{\textbf{es-tar}   } & Ground Truth  & \colorbox{blue!6}{Tamó e’perélachi Concord anelíachi besá makói acres nirú a’lí weká namúti jákami shi.} \\ 
 & Prediction  & \colorbox{green!10}{Concord anelíachi benéalachi, bilé mili akí weká nirú, wekabé namuti nirú.}\\

\hline
\toprule

\end{tabular}}
\end{adjustbox}
    \caption{Example of ground truth and prediction of the Spanish sentence ``Nuestra casa en Concord tiene 30000 acres y un montón de animales." (\textbf{Eng.} \textit{Our home in Concord has 30,000 acres and lots of animals.}) by mBART50. The `z' in `yzuî' of es-oto is actually a Unicode character of code point U+0225 which is a `z' with hook. \colorbox{blue!6}{\textbf{Light blue}: Ground Truth}.  \colorbox{green!10}{\textbf{Light green}: Prediction}.}
    
    \label{tab:ground_truth_prediction}
\end{table*}

\noindent\textbf{mBART50.} For our first multilingual scenario, we fine-tune mBART50 on Spanish-Indigenous data  immediately after tokenization. Similar to our bilingual models, we fine-tune the mBART50 model for $60,000$ steps, measuring performance on Dev every $1,000$ steps, and taking the checkpoint with the best validation loss as our final model used for prediction on Test.

\noindent\textbf{mBART50\textsubscript{curr}.} For the second scenario, mBART50\textsubscript{curr} is first fine-tuned on \texttt{es-en} data for $300$ steps. The validation is done every 20 steps where the checkpoint with lowest loss will be fine-tuned on Spanish-Indigenous language pair for $60,000$ steps, validated every $1000$ steps to pick the best checkpoint with lowest validation loss. Our mBart50\textsubscript{curr} is inspired by the concept of \textit{curriculum learning}~\cite{soviany2021curriculum} where a model can possibly be improved when first trained on an easier task and followed by training on a harder task. In our case here, translating Spanish to English is considered an easier task because mBART50 is pretrained with \texttt{es-en} language pair; whereas Spanish to South American Indigenous languages is considered a more difficult job since mBART50 has not seen any of the $10$ Indigenous languages before.

\section{Results}\label{sec:results}

We evaluate the translation performance with two automatic MT evaluation metrics: BLEU~\cite{papineni-etal-2002-bleu} and chrF~\cite{popovic-2015-chrf}. chrF is an automatic evaluation metric for MT task which can be seen as a F-score for text and has value between $0$ and $1$. BLEU and chrF are the two metrics adopted by AmericasNLP 2021 Shared Task. We surpassed the winner of AmericasNLP2021~\cite{vazquez-etal-2021-helsinki}, in either or both metrics, for $5$ language pairs with the following languages as target: Bribri (bzd), Asháninka (cni), Wixarika (hch), Nahuatl (nah), and Hñähñu (oto). Notably, we double the performance in BLEU score for \texttt{es-bzd}, increasing by about $7.7$ BLEU scores and $0.1$ chrF. We increase $\sim2$ BLEU points in \texttt{es-oto} and $\sim1$ BLEU points in es-hch. For both \texttt{es-cni} and \texttt{es-nah}, we slightly surpass their performance in both metrics. The performance of experiments are shown in Table~\ref{tab:model_results_ninty_dev}. We also offer example predictions in Table~\ref{tab:ground_truth_prediction}.

All surpassing results are achieved by mBART50 or mBART50\textsubscript{curr}. Surprisingly, mBART50\textsubscript{curr} does not consistently improve the performance if compared to mBART50; some of the best performances are achieved by mBART50 (\texttt{es-bzd}, \texttt{es-hch}, \texttt{es-oto}). Nevertheless, mBART50\textsubscript{curr}  performs slightly better than mBART50 on average by $0.076$ BLEU and $0.0034$ chrF. Averagely, mBART50\textsubscript{curr} achieves $7.143$ BLEU score and $0.3134$ chrF while mBART50 achieves $7.068$ BLEU score and $0.31$ chrF. Generally, multilingual models perform better than bilingual model despite that in some language pairs, \texttt{es-en} model performs nearly as good as multilingual models and outperform multilingual models in \texttt{es-aym} and \texttt{es-tar}. For $3$ bilingual models, \texttt{es-en} model generally outperforms the other two  \texttt{es-ca} and \texttt{es-ro} models.

\section{Discussion}\label{sec:discussion}
\subsection{Comparisons to SOTA}
We are able to surpass previous SOTA in five language pairs and  mBART50\textsubscript{curr} achieves $7.143$ BLEU and $0.3134$ chrF on average, comparing to previous SOTA having $6.693$ BLEU and $0.3171$ chrF on average. It can be hypothesized that the reason why we are able to improve average BLEU score by $0.45$,  accomplish comparable average chrF, and surpass in five language pairs is because we use an MT model pretrained on $50$ languages, while~\citet{vazquez-etal-2021-helsinki} pretrain their model only on \texttt{es-en}. We suspect that there could be some languages, other than Spanish and English, which contribute to positive transfer to Indigenous languages. Unlike~\citet{vazquez-etal-2021-helsinki}, we do not leverage external data to build a larger train set. Nor do we build a single multilingual model for all $10$ language pairs, but we rather train one model for each language pair (where every single language pair is independent from the other pairs). The approach of~\citet{vazquez-etal-2021-helsinki} may be able to afford some positive transfer between different Indigenous languages, and hence can be one of our future directions.

\subsection{Fusional to Polysynthetic Translation}\label{sec:poly_to_fusional}

There is literature showing that when translating between a polynthetic\footnote{Polysynthetic languages generally have a more complex morphological system, possibly each word consisting of several morphemes~\cite{haspelmath2013understanding, campbell2012indigenous}.} and a fusional language, some morphological information of the polysynthetic language is `lost'. This is especially relevant to our work since Spanish is a fusional language and many Indigenous languages in our work are polysynthetic~\cite{mager-etal-2021-findings}. \citet{DBLP:journals/corr/abs-1807-00286} carry out a morpheme-to-morpheme alignment between Spanish and polynthetic Indigenous languages, including Nahuatl (nah) and Wixarika (hch) which are both in our data and show that the meanings carried by some  polysynthetic morphemes have no Spanish counterpart. This makes it difficult to translate from polysynthetic languages to fusional Spanish without losing some morphological information. This is also a challenge to translate from fusional Spanish to polysynthetic languages, as there may be no contexts provided to infer the missing parts. This is particularly the case for sentence-level (vs. document level) translation. 

We hypothesize that if there is loss in morphological information when translating from a fusional to polysynthetic languages, either or both the sentence length and word length of prediction will be shorter than the gold standard because some parts in the prediction are left out while the ground truth may contain them. We therefore compare average sentence length and average word length between our gold standard and prediction as shown in Table~\ref{tab:avg_sent_word_len}. However, we find that this hypothesis does not hold for most language pairs as most of them are having similar average sentence and word lengths in gold standard and predictions. We suspect that this is because the test sets are translated from Spanish to Indigenous languages by human translators in a sentence-level fashion, the translators may leave out the missing morphological information when translating Spanish into Indigenous languages due to inability to infer the missing information. As~\citet{DBLP:journals/corr/abs-1807-00286} state, \blockquote{The important Wixarika independent asserters “p+” and “p” are the most frequent morphemes in this language. However, as they have no direct equivalent in Spanish, their translation is mostly ignored. \dots This is particularly problematic for the translation in the other direction, i.e., from Spanish into Wixarika, as a translator has no information about how the target language should realize such constructions. Human translators can, in some cases, infer the missing information. However, without context it is generally complicated to get the right translation.} As this is a sentence-level translation task where contexts can be hard to infer, the gold standard may not contain these parts at the first place. However, a further qualitative linguistic investigation is required to spot the cause of this phenomenon.

\begin{table}[]
\tiny 
\centering
 \begin{tabular}{
    c
    c
    c
    c
    c
    }
 \toprule
\textbf{Target} &
\textbf{Sent (Gold)} & 
\textbf{Sent (Pred)}  &
\textbf{Word (Gold)} &
\textbf{Word (Pred)}   \\
\toprule
aym  & 6.71 & 7.97  & 7.83 & 5.88  \\

bzd  & 11.66 & 10.83  & 3.79 & 3.86 \\

cni  & 6.41 & 6.1  & 8.57 & 8.17 \\

gn   & 6.46 & 6.66  & 6.5 & 6.46 \\

hch  & 9.97 & 8.55 &  5.35 & 5.61  \\

nah  & 6.7 & 6.9 &  7.11 & 7.16 \\

oto  & 10.38 & 9.69  & 4.47 & 4.01  \\

quy  & 6.73 & 6.04  & 7.71 & 8.19  \\

shp  & 8.82 & 7.77 & 5.95 & 5.98   \\

tar  & 9.36 & 8.75  & 5.15 & 4.86  \\

\toprule 
\end{tabular}
\caption{The averages of sentence and word length of test set. The predictions are produced by mBART50. Sent (Gold) and Sent (Pred) are the average sentence length of gold standard and prediction, respectively. Word (Gold) and Word (Pred) are the average word length of gold standard and prediction, respectively. Sentence length is calculated as number of words in each sentence (by splitting sentence with whitespace). Word length is calculated as number of characters in each word.}  \label{tab:avg_sent_word_len}

\end{table}
\normalsize

\subsection{Tokenization with Parent Vocabulary}\label{target_side_subword_issue}
As discussed in Section~\ref{data_preprocessing}, we re-use the tokenizer of parent models without building new ones for child language pairs. We observe that the tokens in target sentences tend to be very short. That is, tokens in these target sentences often consist of one or two characters as can be seen in Table~\ref{tab:tokenization-result}. Hence, target sentences do seem to be encountering oversegmentation. This could be causing loss of meaning as these smaller segments differ from what would be suited for a given Indigenous language.

We further offer statistics related to tokenization with the calculation details provided in Appendix~\ref{appendix:tokenization_stats}, and results shown in Table~\ref{tab:token_stats_ninty_percent_dev} (in Appendix). The difference between the average length of tokens in source and target languages is quite large. For example, for the language pair \texttt{es-bzd}, when tokenized with the \texttt{es-en} tokenizer, average token length for the source language is $3.43$ while that for the target language is $1.21$. This indicates that tokens in source data consist averagely of $\sim3.5$ characters while tokens in target data consist averagely of $\sim1.2$ characters. For this particular \texttt{es-bzd} language pair whose words in target sentences are on average oversegmented into nearly one character per token, the performance is surprisingly better than the previous SOTA. For the other nine language pairs whose words in target sentences are segmented into tokens consisting of $\sim1$ to $\sim2$ characters, the models are still capable of reasonably carrying out the translation task. As~\citet{kocmi-bojar-2020-efficiently} conjecture, this may be a case in point where a model is able to simply generalize well to short subwords. 

\subsection{Non-Standard Orthography}
 Based on a pilot investigation, we find the lack of orthographic standardization to be potentially problematic. We place relevant sample predictions in Table~\ref{tab:ground_truth_prediction}. For example, for the prediction of \texttt{es-aym} pair, we find that a word is predicted nearly correctly with just a difference in one character: ground truth `ukhamarac' is predicted to be `ukhamarak'. As~\citet{coler2014grammar} point out, this may be an issue of non-standard orthography since some Aymara speakers do not consistently differentiate between `c' and `k'. It can be hypothesized that the model generalizes to the `ukhamarak' as a translation of a phrase/word because of potentially relatively higher number of occurrences of `ukhamarak' than `ukhamarac' in training data. In fact, 'ukhamarak' (including its variants with  characters following such as in 'ukhamaraki' and `ukhamarakiw') appears $489$ times in the training set while `ukhamarac' appears zero time (it only exist in test set). Although  'ukhamarac' and 'ukhamarak' can be viewed as the  same word, these are still not counted as a match by some automatic evaluation metrics (including BLEU metrics which we adopt in this work). Interestingly, cases such as the current one illustrates a challenge for automatic MT metrics when evaluating on languages without standard orthography.

\section{Conclusion} \label{sec:conclusion}
In this paper, we describe how we apply transfer learning to MT from Spanish to ten low-resource South American Indigenous languages. We fine-tune pretrained bilingual and multilingual MT models on downstream Spanish to Indigenous language pairs and show the utility of these models. We are able to surpass SOTA in five language pairs using multilingual pretrained MT models without leveraging any external data. Empirically, our results show that this method performs robustly even with an oversegmentation issue on the target side. We also discussed multiple issues that interact with our task, including translating between languages of different morphological structures, effect of tokenization, and non-standard orthography. %

\section*{Acknowledgements}
We appreciate Dr. Miikka Silfverberg for the useful  discussions and the support from the Natural Sciences and Engineering Research Council of Canada, the Social Sciences and Humanities Research Council of Canada, Canadian Foundation for Innovation, Compute Canada (\url{www.computecanada.ca}), and Advanced Research  Computing at University of British Columbia  (\url{https://doi.org/10.14288/SOCKEYE}).
\bibliography{anthology,custom}

\appendix

\section{Appendix} \label{sec:appendix}
\subsection{Additional Experiments}\label{appendix:additional_exp}
We conduct additional experiments for Track Two as mentioned in Section~\ref{sec:datasets}. This additional experiment have identical settings as Track One except that the train set does not involve sentences in development set. We surpass the state-of-the-art performance in $4$ out of $10$ language pairs in either or both BLEU and chrF. Similar to the results in Track One, multilingual MT models perform better than bilingual ones while there are no consistent winner between mBART50 and mBART50\textsubscript{curr}.

\begin{table*}[t]
\footnotesize
\centering
 \begin{tabular}{
    c
    c
    S[table-number-alignment = right] 
    S[table-number-alignment = right]
    S[table-number-alignment = right]
    S[table-number-alignment = right]
    L
    L
    L
    L
    }
 \toprule
\multicolumn{1}{L}{\textbf{Model}} &
\multicolumn{1}{c}{\textbf{Target}} &
\multicolumn{1}{L}{\textbf{Dev BLEU}}  & 
\multicolumn{1}{L}{\textbf{Dev chrF}}  & 
\multicolumn{1}{L}{\textbf{Test BLEU}} & 
\multicolumn{1}{L}{\textbf{Test chrF}}  & 
\multicolumn{1}{L}{\textbf{SOTA BLEU}} & 
\multicolumn{1}{L}{\textbf{SOTA chrF}}  \\
\toprule
es-ca & & 2.415 & 0.227 & 1.0 & 0.197\\ 
es-en & \multirow{3}{*}{aym} & 2.503 & 0.261 & 1.253 & 0.22 & \textbf{\multirow{3}{*}{2.29}} & \textbf{\multirow{3}{*}{0.283}} \\ 
es-ro & & 2.642 & 0.2666 & 1.369 & 0.2273 & \\ 
mBART50 & & 3.105 & 0.275 & 1.38 & 0.236\\
mBART50\textsubscript{curr} & & 3.034 & 0.2679 & 1.37 & 0.2291\\
\hdashline
es-ca & & 2.033 & 0.15 & 2.217 & 0.153\\ 
es-en & \multirow{3}{*}{bzd} & 2.987 & 0.168 & 3.437 & 0.178 & \multirow{3}{*}{2.39} & \multirow{3}{*}{0.165} \\ 
es-ro & & 2.803 & 0.1709 & 3.308 & 0.1816\\ 
mBART50 & & 4.205 & 0.188 &  4.272 & \B 0.197\\ 
mBART50\textsubscript{curr} & & 4.072 & 0.1871 & \B 4.438 & 0.1911\\
\hdashline
es-ca & & 2.628 & 0.212 & 2.429 & 0.201\\ 
es-en & \multirow{3}{*}{cni} & 1.671 & 0.212 & 1.623 & 0.208 & \multirow{3}{*}{3.05} & \textbf{\multirow{3}{*}{0.258}}\\ 
es-ro & & 1.639 & 0.2225 & 1.829 & 0.209 & \\ 
mBART50 & & 3.074 & 0.26 & \B 3.539 & 0.25\\ 
mBART50\textsubscript{curr} & & 3.404 & 0.2573 & 3.537 & 0.2491 \\
\hdashline
es-ca & & 3.637 & 0.245 & 3.523 & 0.254\\ 
es-en & \multirow{3}{*}{gn} & 4.206 & 0.282 & 4.217 & 0.297 & \textbf{\multirow{3}{*}{6.13}} & \textbf{\multirow{3}{*}{0.336}}\\ 
es-ro & & 3.784 & 0.2771 & 4.699 & 0.291 & \\ 
mBART50 & & 4.911 & 0.287 & 4.801 & 0.304\\
mBART50\textsubscript{curr} & & 4.496 & 0.2795 & 4.702 & 0.2918 \\
\hdashline
es-ca & & 5.618 & 0.191 & 7.595 & 0.197\\ 
es-en & \multirow{3}{*}{hch} & 6.578 & 0.234 & 8.995 & 0.245 & \multirow{3}{*}{9.63} & \textbf{\multirow{3}{*}{0.304}} \\ 
es-ro & & 7.536 & 0.2594 & 10.123 & 0.2732 & \\ 
mBART50 & & 8.617 & 0.254 & 11.526 & 0.272\\
mBART50\textsubscript{curr} & & 9.067 & 0.2582 & \B 11.539 & 0.2731 \\
\hdashline
es-ca & & 0.753 & 0.239 & 0.705 & 0.222\\ 
es-en & \multirow{3}{*}{nah} & 0.73 & 0.25 & 0.772 & 0.22 & \textbf{\multirow{3}{*}{2.38}} & \textbf{\multirow{3}{*}{0.266}} \\ 
es-ro & & 1.06 & 0.2619 & 0.6983 & 0.2363\\ 
mBART50 & & 1.69 & 0.281 & 1.497 & 0.255\\ 
mBART50\textsubscript{curr} & & 1.704 & 0.2731 &1.78 & 0.2412 \\

\hdashline
es-ca & & 0.536 & 0.122 & 0.86 & 0.12\\ 
es-en & \multirow{3}{*}{oto} & 0.745 & 0.124 & 1.039 & 0.121 & \textbf{\multirow{3}{*}{1.69}} & \textbf{\multirow{3}{*}{0.147}} \\ 
es-ro & & 0.5125 & 0.1198 & 0.8811 & 0.1226\\ 
mBART50 & & 0.816 & 0.133 & 1.354 & 0.132\\
mBART50\textsubscript{curr} & & 0.8851 &0.1348 &1.338 &0.1331 \\
\hdashline
es-ca & & 2.199 & 0.322 & 2.191 & 0.328\\ 
es-en & \multirow{3}{*}{quy} & 2.217 & 0.337 & 2.892 & 0.347 & \multirow{3}{*}{2.91} & \multirow{3}{*}{0.346} \\ 
es-ro & & 2.081 & 0.3416 & 2.094 & 0.3539 & \\ 
mBART50 & & 2.242 & 0.356 & \B 3.167 & \B 0.366 &  \\ 
mBART50\textsubscript{curr} & & 2.516 &0.355 &3.038 &0.3659\\

\hdashline
es-ca & & 1.511 & 0.178 & 1.234 & 0.168\\ 
es-en & \multirow{3}{*}{shp} & 2.134 & 0.21 & 2.017 & 0.196 & \textbf{\multirow{3}{*}{5.43}} & \textbf{\multirow{3}{*}{0.329}} \\ 
es-ro & & 1.964 & 0.2205 & 1.43 & 0.2048 & \\ 
mBART50 & & 2.131 & 0.194 & 2.013 & 0.185\\ 
mBART50\textsubscript{curr} & & 2.067 &0.1947 &1.809 &0.1856 \\

\hdashline
es-ca & & 0.256 & 0.095 & 0.047 & 0.084\\ 
es-en & \multirow{3}{*}{tar} & 0.034 & 0.057 & 0.023 & 0.05 & \textbf{\multirow{3}{*}{1.07}} & \textbf{\multirow{3}{*}{0.184}} \\ 
es-ro & & 0.1583 & 0.09438 & 0.2985 & 0.08932\\ 
mBART50 & & 0.09 & 0.093 & 0.073 & 0.101\\ 
mBART50\textsubscript{curr} & & 0.1212 &0.09463 &0.09013 &0.1007 \\

\toprule 
\end{tabular}
\caption{Modeling results of Track Two. The boldfaced numeric values are the best performances. SOTA values represent the state-of-the-art performance which are all from~\citet{vazquez-etal-2021-helsinki} except that the es-quy SOTA chrF value is from~\cite{moreno-2021-repu}. Source language is always Spanish so it is ignored.}  \label{tab:model_results_no_dev}

\end{table*}
\normalsize

\subsection{Tokenization Output}\label{appendix:tokenization_stats}
As mentioned in Section~\ref{target_side_subword_issue}, we calculate statistics related to tokenization on training data as shown in Table~\ref{tab:token_stats_ninty_percent_dev}. To calculate these statistics, padding tokens, end of sentence tokens and the underscore (or more precisely, U+2581) prepended due to sentencePiece technique~\cite{kudo-richardson-2018-sentencepiece} are removed from the tokenized sentences. Sentence length is calculated as number of tokens in a sentence. Token length is calculated as the number of characters in a token. Average sentence length is calculated by averaging the sentence lengths of all sentences. Average token length is calculated as 
$$\frac{\sum_{i=1}^N\sum_{j=1}^{n_i}{|s_{ij}|}}{\sum_{i=1}^N{n_i}}$$ where $n_i$ denotes the number of tokens in $i^{th}$ sentence and $N$ denotes the number of sentences in training data. $|s_{ij}|$ denotes the length (number of characters) of $j^{th}$ token in $i^{th}$ sentence.

\begin{table*}[t]
\footnotesize
\centering
 \begin{tabular}{
    c
    c
    S[table-number-alignment = right, table-format=2.2]
    S[table-number-alignment = right, table-format=2.2]
    S[table-number-alignment = right, table-format=2.2]
    S[table-number-alignment = right, table-format=1.2]
    }
 \toprule
\multicolumn{1}{L}{\textbf{model}} &
\multicolumn{1}{L}{\textbf{Target Lang}} & 
\multicolumn{1}{L}{\textbf{source avg sentence length}} & \multicolumn{1}{L}{\textbf{target avg sentence length}} & \multicolumn{1}{L}{\textbf{source avg token length}}  & \multicolumn{1}{L}{\textbf{target avg token length}} \\
\toprule

es-ca &   & 26.37 & 49.1 & 3.61 & 1.81 \\
es-en & \multirow{2}{*}{aym}  & 24.55 & 45.07 & 3.88 & 1.99 \\
es-ro &   & 25.74 & 47.9 & 3.71 & 1.91 \\
mBART50 &  & 27.4 & 37.85 & 3.66 & 2.53 \\
\hdashline
es-ca &  & 9.42 & 22.42 & 3.24 & 1.28 \\
es-en & \multirow{2}{*}{bzd}  & 8.9 & 21.43 & 3.43 & 1.21 \\
es-ro &   & 9.13 & 21.52 & 3.34 & 1.23 \\
mBART50 &  & 10.75 & 19.67 & 3.3 & 1.54 \\
\hdashline
es-ca &  & 17.6 & 30.56 & 3.33 & 1.92 \\
es-en & \multirow{2}{*}{cni}  & 16.72 & 27.78 & 3.51 & 2.12 \\
es-ro &  & 17.31 & 29.17 & 3.44 & 2.04 \\
mBART50 &  & 19.38 & 23.9 & 3.27 & 2.69 \\
\hdashline
es-ca &   & 31.89 & 50.6 & 3.69 & 2.01 \\
es-en & \multirow{2}{*}{gn}  & 30.15 & 50.77 & 3.9 & 2.0 \\
es-ro &   & 31.92 & 52.45 & 3.73 & 1.97 \\
mBART50 &   & 33.79 & 41.34 & 3.63 & 2.6 \\
\hdashline
es-ca &   & 11.15 & 23.01 & 3.24 & 1.68 \\
es-en & \multirow{2}{*}{hch}  & 10.49 & 21.56 & 3.44 & 1.79 \\
es-ro &   & 10.76 & 22.27 & 3.35 & 1.73 \\
mBART50 &   & 13.34 & 20.14 & 3.08 & 2.17 \\
\hdashline
es-ca &   & 33.7 & 51.39 & 3.03 & 1.83 \\
es-en & \multirow{2}{*}{nah}  & 34.36 & 49.58 & 2.96 & 1.94 \\
es-ro &   & 34.44 & 51.52 & 2.95 & 1.83 \\
mBART50 &  & 36.78 & 45.54 & 2.87 & 2.32 \\
\hdashline
es-ca &   & 18.0 & 37.72 & 3.14 & 1.64 \\
es-en & \multirow{2}{*}{oto}  & 18.2 & 36.06 & 3.1 & 1.51 \\
es-ro &   & 18.49 & 37.58 & 3.07 & 1.7 \\
mBART50 &  & 20.62 & 32.91 & 2.98 & 1.82 \\
\hdashline
es-ca &   & 20.16 & 42.8 & 3.65 & 1.83 \\
es-en & \multirow{2}{*}{quy}  & 19.26 & 37.68 & 3.82 & 2.08 \\
es-ro &   & 20.16 & 41.45 & 3.73 & 1.92 \\
mBART50 &   & 22.96 & 31.47 & 3.42 & 2.65 \\
\hdashline
es-ca &   & 9.71 & 16.53 & 3.19 & 1.75 \\
es-en & \multirow{2}{*}{shp} & 9.06 & 15.56 & 3.41 & 1.85 \\
es-ro &   & 9.42 & 15.84 & 3.28 & 1.82 \\
mBART50 &   & 11.12 & 13.54 & 3.23 & 2.5 \\
\hdashline
es-ca &   & 12.48 & 19.4 & 2.97 & 1.48 \\
es-en & \multirow{2}{*}{tar}  & 12.83 & 18.32 & 2.89 & 1.57 \\
es-ro &   & 13.08 & 19.33 & 2.84 & 1.5 \\
mBART50 &  & 14.15 & 15.64 & 2.98 & 2.16 \\

\toprule 
\end{tabular}
\caption{Token statistics for our Train set. The way of calculating these  figures is presented in Appendix~\ref{appendix:tokenization_stats}. Since mBART50 and mBART50\textsubscript{curr} are having exactly same statistics as they use same tokenizer, the statistics of mBART50\textsubscript{curr} are ignored.}  \label{tab:token_stats_ninty_percent_dev}

\end{table*}
\normalsize

\end{document}